\documentclass[10pt,twocolumn,letterpaper]{article}

\usepackage{iccv}
\usepackage{times}
\usepackage{epsfig}
\usepackage{graphicx}
\usepackage{amsmath}
\usepackage{amssymb}
\usepackage{pifont}
\usepackage{soul}
\usepackage[dvipsnames, table]{xcolor}
\usepackage{float}
\usepackage{caption}
\usepackage{subcaption}
\usepackage{booktabs}
\usepackage{makecell}
\usepackage{listings}

\usepackage{pifont}

\usepackage{multirow}
\usepackage[breaklinks=true,bookmarks=false]{hyperref}

\iccvfinalcopy 

\definecolor{codegreen}{rgb}{0,0.6,0}
\definecolor{codegray}{rgb}{0.5,0.5,0.5}
\definecolor{codepurple}{rgb}{0.58,0,0.82}
\definecolor{backcolour}{rgb}{0.95,0.95,0.92}

\lstdefinestyle{mystyle}{
    backgroundcolor=\color{backcolour},   
    commentstyle=\color{codegreen},
    keywordstyle=\color{magenta},
    numberstyle=\tiny\color{codegray},
    stringstyle=\color{codepurple},
    basicstyle=\ttfamily\footnotesize,
    breakatwhitespace=false,         
    breaklines=true,                 
    captionpos=b,                    
    keepspaces=true,                 
    numbers=left,                    
    numbersep=5pt,                  
    showspaces=false,                
    showstringspaces=false,
    showtabs=false,                  
    tabsize=2
}

\newcommand{\xmark}{\ding{55}}%
\newcommand{\cmark}{\ding{51}}%

\ificcvfinal\fi

\def\ours{\mbox{Le-RNR-Map}\xspace}

\newcommand{\rnr}{\mbox{RNR-Map}\xspace}

\newcommand\blfootnote[1]{%
  \begingroup
  \renewcommand\thefootnote{}\footnote{#1}%
  \addtocounter{footnote}{-1}%
  \endgroup
}
\makeatletter

\def\blfootnote{\gdef\@thefnmark{}\@footnotetext}
\makeatother

\begin{document}

\title{Language-enhanced RNR-Map: Querying  Renderable Neural Radiance Field maps with natural language\vspace{-0.4cm}}

\author{
Francesco Taioli\textsuperscript{1,*}, Federico Cunico\textsuperscript{1,*}, Federico Girella\textsuperscript{2,*},\\ Riccardo Bologna\textsuperscript{2,*,\textdagger},
Alessandro Farinelli\textsuperscript{2}, Marco Cristani\textsuperscript{1} \and 
{\small \textsuperscript{1}University of Verona, Dept. of Engineering for Innovation Medicine}\\
{\small \textsuperscript{2}University of Verona, Dept. of Computer Science} \\
{\tt\small name.surname@univr.it} , {\tt\small name.surname@studenti.univr.it\textsuperscript{\textdagger}}
}

\ificcvfinal\thispagestyle{empty}\fi

\twocolumn[{
\maketitle
\vspace{-0.9cm}
\begin{center}
    \captionsetup{type=figure}
    \includegraphics[width=0.7\linewidth]{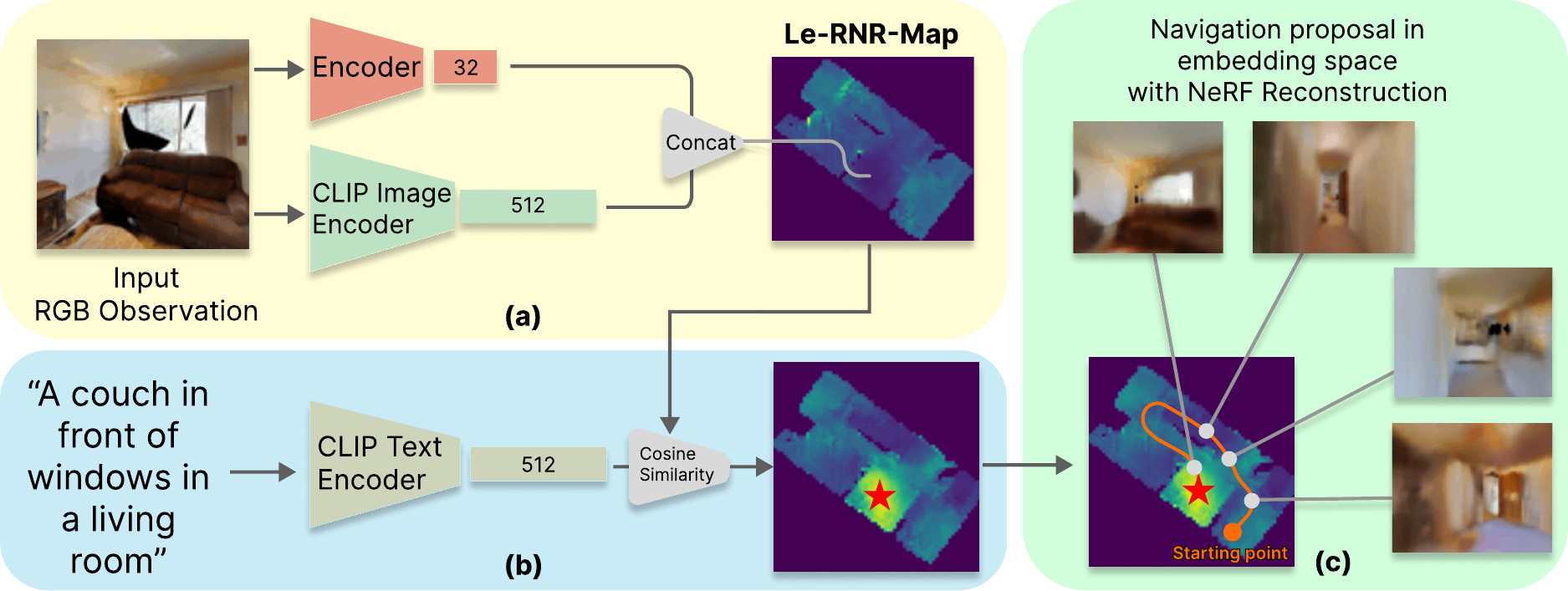}
    \caption{Overview of \ours. (a) shows \ours map construction in which we embed  visual and visual-language-aligned features, (b) shows the query search, at inference time, with natural language, and (c) shows the reconstruction of the images on the path, from starting position (orange circle) to the goal (red star) using NeRF.}
    \label{fig:teaser}
\end{center}
}]

\begin{abstract}

We present \ours, a Language-enhanced Renderable Neural Radiance map for Visual Navigation with natural language query prompts. The recently proposed \rnr employs a grid structure comprising latent codes positioned at each pixel.
These latent codes, which are derived from image observation, enable: \textit{i)} image rendering given a camera pose, since they are converted to Neural Radiance Field; \textit{ii)} image navigation and localization with astonishing accuracy. 
On top of this, we enhance \rnr with CLIP-based embedding latent codes, 
allowing natural language search without additional label data.
We evaluate the effectiveness of this map in single and multi-object searches. We also investigate its compatibility with a Large Language Model as an ``affordance query resolver''.
 Code and videos are available at the link \href{https://intelligolabs.github.io/Le-RNR-Map/}{https://intelligolabs.github.io/Le-RNR-Map/}.
\end{abstract}

\vspace{-0.5cm}
\section{Introduction}

\blfootnote{\textsuperscript{*}The authors contributed equally to this paper}
Embodied AI is receiving a lot of attention in recent years, with interesting yet very challenging tasks such as Embodied Question Answering \cite{eqa_matterport}, Image-Goal Navigation \cite{krantz2022instancespecific}, Visual-Language Navigation \cite{krantz2020navgraph} and zero-shot Object-Goal navigation \cite{majumdar2022zson,gadre2022cows}.
Currently, some areas of research are focusing on creating explicit map representations that could improve the performances on those tasks, such as semantic segmentation maps \cite{chaplot2020object,marza2023autonerf}, occupancy maps and top-down semantic map prediction \cite{georgakis2022cm2}. More recently, some works attempted to embed latent vectors in the explicit map to perform navigation \cite{visualLanguageMap}.
However, very few tried to combine Visual-Language-aligned and NeRF \cite{mildenhall2020nerf} latent codes into the map itself to solve different tasks simultaneously.

For this reason, in this preliminary work, we expand \rnr \cite{rnrMap}
by enhancing it with visual-language-aligned features.
The result is a \textit{Language enhanced, Renderable Neural Radiance Map} (\ours) for visual navigation that is \textit{visually descriptive}, thanks to the Neural Radiance Field embeddings, \textit{generalizable}, since it uses off-the-shelf models requiring no training, and queryable with \textit{both text and images} thanks to the addition of visual-language-aligned features.
To the best of our knowledge, this is the first work in the literature to create a map representation that allows to solve three distinct tasks \textit{at the same time}: \textit{i)}~Localization of objects given a query image, by using the \rnr \cite{rnrMap} embeddings; \textit{ii)}~Open Vocabulary localization of objects through natural language, using Visual-Language-aligned embeddings; \textit{iii)}~rendering of the path to the target, highlighting the searched object with Visual-Language-aligned features, without the need to physically move the agent.

\section{Related Literature}

\subsection{Maps For Navigation}
There is an extensive research area that focuses on how to build maps to aid navigation in indoor environments
\cite{chaplot2020learning,chen2019learning,marza2023autonerf,visualLanguageMap,rnrMap,taioli2023unsupervised}. An occupancy grid map is a $m \in \mathbb{R}^{M \times M \times C}$ matrix, where $M \times M$ is the spatial size and $C$ is the number of channels storing information about a corresponding region.
Recently, the authors of \rnr ~\cite{rnrMap} introduced a novel type of map, in which the latent codes are embedded from visual observations and can be converted to a neural radiance field, which enables image rendering given a camera pose, thus being visually descriptive. Moreover, the author showed that this novel type of map can be useful for visual localization and navigation. 

In AutoNeRF \cite{marza2023autonerf}, the authors introduce a method to collect data required to train NeRFs using autonomous embodied agents, and use the experience to build an implicit map representation of the environment. Moreover, they augment the NeRF rendering procedure with a segmentation head over $S$ predefined classes. 
In VLMaps \cite{visualLanguageMap}, the authors ground language information to visual observation 
fusing pre-trained visual language features \cite{li2022languagedriven} into a 2D spatial representation.
A similar approach is presented in
\cite{chen2022nlmapsaycan}. In contrast to \cite{visualLanguageMap}, \ours also allows us to perform Image-Goal Navigation, and NeRF rendering given a camera pose. Additionally, \ours can be built faster, in around 60 seconds, for $\sim$1000 RGB-D images ($128\times 128$). 

\subsection{Nerf}

 Neural Radiance Fields, introduced in \cite{mildenhall2020nerf}, address the problem of view synthesis, that is generating scene views from unseen novel viewpoints.
 NeRF scenes are modelled by a multilayer perceptron network which outputs the radiance emitted by a 3D point, given a spatial location $(x,y,z)$ and a viewing direction ($\theta$, $\phi$). The original NeRF formulation \cite{mildenhall2020nerf} can only represent small scenes, and does not generalize to new scenes/objects.
 To solve these limitations, other works take the challenge of learning a distribution over complex scenes. Generative Scene Networks (GSNs) \cite{gsn} can be used to learn a rich scene prior in order to generate new scenes or fill the given one, decomposing the scene in local radiance fields that can be rendered from a moving camera. Moreover, they can be used to render images from latent codes, which in our case are stored in \ours, like the original \rnr formulation \cite{rnrMap}.
 Some recent works have extended the principle goal of NeRF with some other tasks. LERF \cite{kerr2023lerf} proposed a method for grounding CLIP representations in a dense, multi-scale 3D field, which can render dense relevancy maps given textual queries. However, LERF is still limited to small scenes and requires 45 minutes for a capture.

\subsection{Language and Vision}
Mapping text and images is the problem of estimating a function that maps images to the desired text (e.g. captioning \cite{luo2022semanticconditional}) and vice versa (e.g image generation with text-conditioning \cite{rombach2022highresolution,saharia2022photorealistic}). Introduced in \cite{clip}, CLIP is a model composed of an image-encoder and text-encoder trained to map embeddings from images and their description close to each other in the feature space, with a contrastive loss that enforces non-related pairs to be mapped further from each other.
The authors showed with extensive experiments that CLIP achieves competitive zero-shot performance and thus can be used as a foundational model in a variety of task, such as scene segmentation \cite{peng2023openscene,li2022languagedriven}, Open-Vocabulary object detection \cite{gu2022openvocabulary} and Image-Generation \cite{rombach2022highresolution}.
Moreover, \cite{dong2023maskclip} introduces MaskCLIP, a 
framework for obtaining scene segmentation with indirect supervision from language.
\section{Method}

\subsection{Map creation}
We create a \ours by first extracting visual and Visual-Language-aligned features from RGB frames, 
while a subsequent feature registration process projects them in the map (Fig. \ref{fig:teaser}a).

\textbf{Visual embeddings.}
Inspired by \cite{rnrMap} we consider a robot agent exploring the scene with a random walk, using RGB-D data and its on-board sensors, \ie{} odometry information, to build \ours.
An encoder-decoder architecture performs the creation of the \rnr.
To allow an effective encoding of the 3D environment, the authors of \cite{rnrMap} perform training of the encoder-decoder as follows: \textit{i)} the encoder takes in input the RGB-D image and extracts the pixel features; \textit{ii)} the decoder uses pixel features, along with the current pose (\ie{} position of the agent), to sample latent information along each camera ray corresponding to each pixel, and tries to render the corresponding images.
The latent codes extracted from the encoder, then, represent the pixel-level visual information from the current view. In our experiments, the encoding features are $F_{rnr} \in \mathbb{R}^{32}$. These features allow image rendering using Neural Radiance Field, and image localization in the map. 
For a more in-depth description, we refer the reader to \cite{rnrMap}.

\textbf{Natural language.} To include language features, we use a pre-trained CLIP~\cite{clip} image-encoder to get $F_{clip} \in \mathbb{R}^{512}$ from the current RGB-D frame.

\textbf{\ours}. The final embedding space for the current observation RGB-D frame is then composed as $F_{le-rnr} = F_{clip} \oplus F_{rnr}$. $F_{le-rnr} \in \mathbb{R}^{544}$ with the first 32 channels generated from \rnr and the 512 remaining from CLIP.
Both the features from RNR and CLIP are then projected to the 2D map using the depth information, as in \cite{rnrMap}.
This allows us to keep the exact performances of \rnr for the Image-Goal navigation task, with the addition of being able to query the navigation through natural language thanks to the CLIP features.

\subsection{Language-vision object search}\label{sec:negative_prompts}
\begin{figure}[t]
    \centering
    \includegraphics[width=\linewidth]{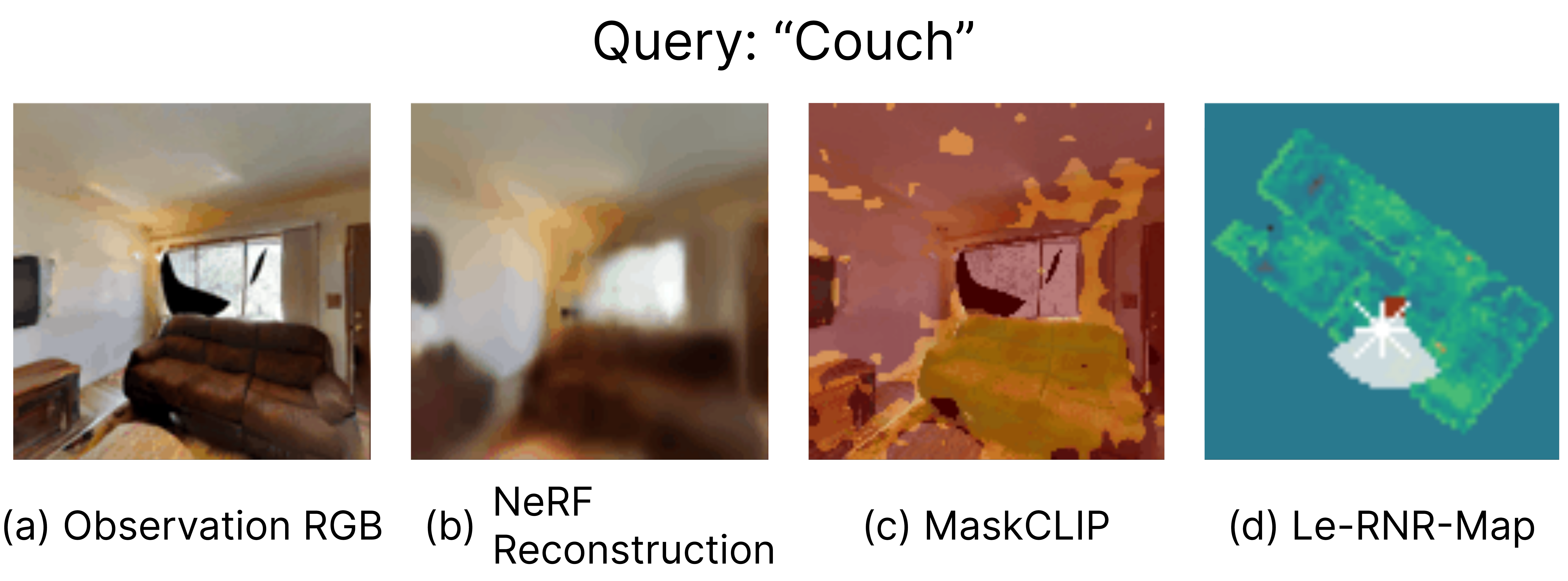}
    \caption{(a) Observation from Habitat-sim \cite{habitat19iccv}. (b) Reconstruction using latent code from \ours using Neural Radiance Field (c) Feature visualization of the text query using MaskCLIP \cite{dong2023maskclip}. (d) Top-down view of \ours. }
    \label{fig:query}
    \vspace{-0.5cm}
\end{figure}

The vision-language object search is performed by providing a natural language query (Fig. \ref{fig:teaser}b). This query should indicate the objects required to find (either big furniture or small objects) that the navigation module has to handle as goal objects. 
Once the query is provided, the text embeddings are extracted with a pre-trained CLIP~\cite{clip} text-encoder, and the cosine similarity with each cell of the \ours is computed. We select the location of maximal similarity as the goal location.
The 3D end-goal predicted location, given the $(x, y)$ indices expressed in \ours coordinate system, is obtained through inverse projection as in \cite{rnrMap}. The exact procedure is presented in the supplementary material.
To ensure the correct maximal similarity is found, prompt engineering is performed with negative prompting following \cite{kerr2023lerf,clip}. Together with the query prompt, this empirically shows a more fine-grained similarity on the maps and gives better localization as shown in Fig. \ref{fig:similarity_maps}.
An extension of GSN~\cite{gsn} is then used to synthesize novel views and render a possible pathway that leads from any point starting point of the map to the required goal. Once the navigation reaches the proximity of the goal (\ie{} the location of maximal similarity in the \ours), we estimate the camera orientation towards the goal by rotating the camera by $360^\circ$, computing the CLIP features for each degree and choosing the one with maximal cosine similarity with the target query.
Finally, the visual saliency of the goal (Fig. \ref{fig:query}c) is extracted from the RGB reconstructed by GSN~\cite{gsn} using pixel-wise CLIP features~\cite{dong2023maskclip}.
When multiple target objects are requested, the navigation is performed sequentially for each object following the same procedure, using as starting location the previous target location. 

\section{Experiments}

To highlight the benefits that \ours brings to the Object-Goal Navigation task, we designed a set of targeted experiments.
In Sec.~\ref{subsec:exp_lang_search} we evaluate the ability of \ours to correctly locate items using only Natural Language prompts. In Sec.~\ref{subsec:solv_ambig} we show that the Renderable Neural Radiance map allows the user to properly select the correct item of interest in case of ambiguity. 
Finally, in Sec.~\ref{subsec:exp_afford_search}, we explore the possible collaboration between a Large Language Model and \ours to find items and locations based on contextual prompts, called \textit{Affordance queries} (\eg{} the query \textit{``Find me a drink to wake me up''} results in the agent looking for a cup of coffee).
All of our experiments were conducted inside the Habitat-Sim \cite{habitat19iccv} using the Gibson \cite{xia2018gibson} dataset.

\subsection{Searching items by Language prompts}
\label{subsec:exp_lang_search}
The goal of \ours is to provide the user with a Natural Language interface with the agent. Such an interface would allow the user to prompt the agent with low effort, even in a constrained environment where traditional interactions may be unavailable (\eg{} the user is holding something and can't physically interact with the agent) or improbable (\eg{} asking an agent to look for a particular object by showing it a picture of the object itself). With this goal in mind, we test the ability of \ours to locate different common items and/or locations in the environment, using only the CLIP features embedded in the navigation map.

\begin{figure}
    \centering
    \includegraphics[width=0.8\linewidth]{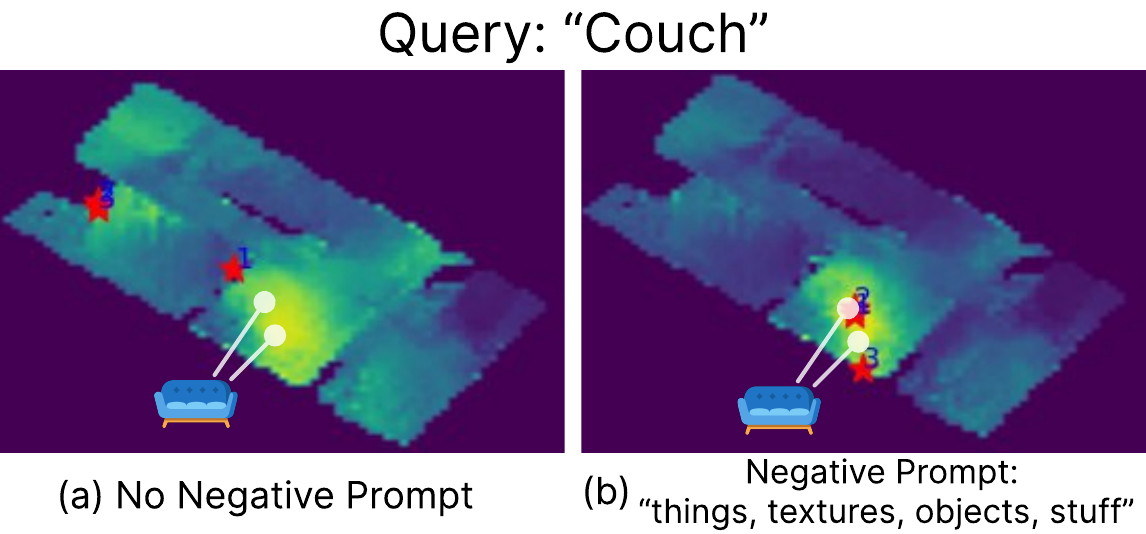}
    \caption{Similarity heatmap between the prompt \texttt{Couch} and \ours. (a) is without negative prompts. (b) shows that negative prompts result in cleaner maps with more concentrated similarity zones. The stars indicate the maximum similarity locations.}
    \label{fig:similarity_maps}
\end{figure}

In Tab. \ref{tab:quantitative} we report the Success Rate and Distance To Success (DTS), as defined in \cite{chaplot2020object}, on some scenes of the validation split of the Gibson tiny dataset \cite{xia2018gibson}.
The dataset provides a textual label for the target object and its location in the scene as ground truth. We use the label as text prompt and compute the metrics comparing our predicted location with the ground truth.
Additionally, as explained in Sec.~\ref{sec:negative_prompts}, we define a series of unwanted objects or general/background elements (\eg{} \texttt{stuff, wall, floor}) as negative prompts. These prompts may be specific for each scene.
Together with the target prompt, we compare their similarity results with the map embeddings (as seen in \cite{clip}) resulting in similarity maps with more consistent areas of interest as shown in Fig.~\ref{fig:similarity_maps}. 
In general, \ours enables us to obtain a decent success rate and DTS  without additional training required. Moreover, we investigate the low Success Rate for the \textit{Darden} scene. We found that the observations, given to the agent during the map creation, contain several artefacts, such as mirror reflections, missing walls and mesh holes, leading to incorrect embeddings into the map. Further study will analyze this problem.

\begin{table}[h]
\begin{center}
\caption{Results on the \texttt{val} split of Gibson tiny dataset. Note that our setup is \textit{known} since we generate the map beforehand. Each scene has a different negative prompt.}
\label{tab:quantitative}
\small
\begin{tabular}{|c|c|c|c|c|}
\hline
\textbf{Scene name} & \makecell{ \textbf{Negative}\\\textbf{Prompts}} & \textbf{Success} $\uparrow$ & \textbf{DTS (m)} $\downarrow$ \\
\hline\hline
\multirow{2}{*}{\textit{Corozal}} & \xmark & 0.69 & 1.19\\
& \cmark & 0.69 & \textbf{0.65} \\
\hline
\multirow{2}{*}{\textit{Darden}} & \xmark & 0.37 & 4.12 \\
& \cmark & \textbf{0.59} & \textbf{2.59} \\
\hline
\multirow{2}{*}{\textit{Markleeville}} & \xmark & 0.81 & \textbf{1.38} \\
& \cmark & \textbf{0.83} & 1.50 \\
\hline
\multirow{2}{*}{\textit{Wiconisco}} & \xmark & 0.76 & 0.50 \\
& \cmark & 0.76 & 0.50 \\
\hline
\hline
\multirow{2}{*}{Average} & \xmark & 0.66 & 1.79 \\
& \cmark & \textbf{0.72} & \textbf{1.31} \\
\hline
\end{tabular}
\end{center}
\vspace{-0.8cm}
\end{table}

\subsection{Solving prompt ambiguities}\label{subsec:solv_ambig}
One of the advantages of having a Renderable Neural Map is the possibility for the agent \textit{to explore the scene without actually moving in the real world.} This is particularly useful in scenarios where the user prompt may be ambiguous and refer to multiple objects or locations in the scene (\eg{} \texttt{window} may refer to different windows). When this happens, the \rnr can be used to provide the user with visual previews of the paths it would take to get to all the possible solutions. In this section, we explore this case and provide a solution using \ours.

First, as in Sec.~\ref{subsec:exp_lang_search}, we compute similarities between the CLIP features extracted from the prompt and the ones embedded in \ours by also using negative prompts. After finding the maximum similarity location, we suppress all the similarities in the adjacent cells of the map. 
We then look for the new maximum similarity above a threshold $th = 0.6$. For each target found, we render the path that the agent would follow to reach the target using a shortest path algorithm. The process ends when there is no longer a similarity score greater than $th$. We consider it a success when this process finds the target item in at least one of the $N$ predicted paths, simulating the user ``selecting'' the desired item. While this evaluation is heavily reliant on hyper-parameters, such as the number $N$ of predicted paths allowed and the value $T$ of the threshold, we still believe it to be an interesting use case, and propose our work as a first informal approach to this problem.
For video examples, we refer the user to the supplementary material. 

\subsection{Affordance search}
\label{subsec:exp_afford_search}

We argue that Natural Language alone is not enough to achieve natural interaction with the user and the agent, especially if we restrict the user to a limited set of words (classes) or a rigid sentence structure. The idea comes from the following observation: what if we want to search for some specific location of an indoor environment, but we are unable to express the query in a direct way? As an example, consider the scenario where we want to search for a location ``\textit{that can be relaxing after a long day at work}''. We define this use case as ``affordance search", and propose to take advantage of the current state-of-the-art Large language model to translate the query and output a set of possible target descriptions, using the available  GPT-3.5 chat-completions API. After retrieving the descriptions, we sequentially search for each target with the procedure presented in Sec. \ref{sec:negative_prompts}.
Visualizations and details about prompts are available in the supplementary material.

\section{Conclusions \& Future works}
In this preliminary study, we have enhanced the novel \rnr \cite{rnrMap} to allow natural language search using an off-the-shelf model. We qualitatively show the result using single and multi-object search, generating videos of the shortest path to reach the object using the neural Radiance Field from latent code inside \rnr. Moreover, we show how LLMs can be used as an ``affordance query resolver''. Several interesting future works could follow this preliminary study. We plan to improve the rendering quality of the reconstructed observation by adding a Language-driven grounding head to the NeRF procedure, similar to \cite{kerr2023lerf} but in an open-scene case.  Research could also focus on training an end-to-end agent to solve the zero-shot object goal navigation, and studying if the online generation of \ours serves as an auxiliary task to improve the generalization capabilities of the agent. 
Furthermore, we plan to evaluate the impact of different negative prompts in the standard zero-shot object goal navigation benchmark.
Also, we plan to study how to deal with dynamic environments, thus providing a way to update the map, both in NeRF and language embedding space.
Finally, we are outlining a real-world implementation in an industrial environment, leveraging Human-Robot-Interaction methods~\cite{cunico2023oo} for better integration of human-guided robot navigation and enhanced task efficiency.

{\small
\bibliographystyle{ieee_fullname}
\bibliography{egbib}
}

\setcounter{section}{0}
\clearpage
{\Large \bf{Appendix}}\\

\section{Video examples}
A video showing examples of our method can be found at \href{https://intelligolabs.github.io/Le-RNR-Map/}{https://intelligolabs.github.io/Le-RNR-Map/}.

\section{Obtaining 3D location given \ours target coordinates}
To retrieve the 3D end-goal location given the $(x, y)$ indices, expressed in \ours coordinate system, we use the following approach: 
first, we define an arbitrary rotation in the \rnr space, which only rotates on the axis perpendicular to the map plane by $\alpha$ radians. We represent this rotation with a $3\times 3$ matrix $R_{\text{2D}}$. We also define a $3\times 1$ vector $t_{\text{2D}}$ as $\begin{bmatrix}x & 0 & y \end{bmatrix}^{T}$, which indicates the translation in the \rnr from the origin.
We can then define the $4\times4$ roto-translation matrix as
\begin{equation}
    R_{t_{\text{2D}}} = \begin{bmatrix} R_{\text{2D}} & t_{\text{2D}} \\ \mathbf{0} & 1 \end{bmatrix}
\end{equation}
where \textbf{0} indicates a $1\times3$ vector of zeros.
Finally, we map the 2D coordinates (up to rotation uncertainty) to the 3D roto-translation matrix $R_{t_{3D}}$ as follows:
\begin{equation}
    R_{t_{\text{3D}}} = R_{t_{\text{origin}}}^{-1} R_{t_{\text{2D}}}
\end{equation}

where $R_{t_{\text{origin}}}$ is the roto-translation matrix of the origin in the 3D system, stored during the map generation process.
$R_{t_{\text{3D}}}$ is structured as follows:
\begin{equation}
    R_{t_{\text{3D}}} = \begin{bmatrix} R_{\text{3D}} & t_{\text{3D}} \\ \mathbf{0} & 1 \end{bmatrix} 
\end{equation}

and $t_{\text{3D}}$ is the desired 3D pose of the agent.

\section{Negative Prompts}
In Tab. \ref{tab:quantitative} we report  the Table of the main paper, in which we add details about the corresponding negative prompts:
\begin{itemize}
    \item \textit{Corozal}: \texttt{"wc"}
    \item \textit{Darden}: \texttt{"the floor inside the house", "the wall inside the house"}
    \item \textit{Markleeville}: \texttt{"the floor inside the house"}
    \item \textit{Wiconisco}:  \texttt{"things", "stuff", "textures", "objects"}
\end{itemize}

\section{Object Search}
\begin{figure}[h]
    \centering
    \includegraphics[width=\linewidth]{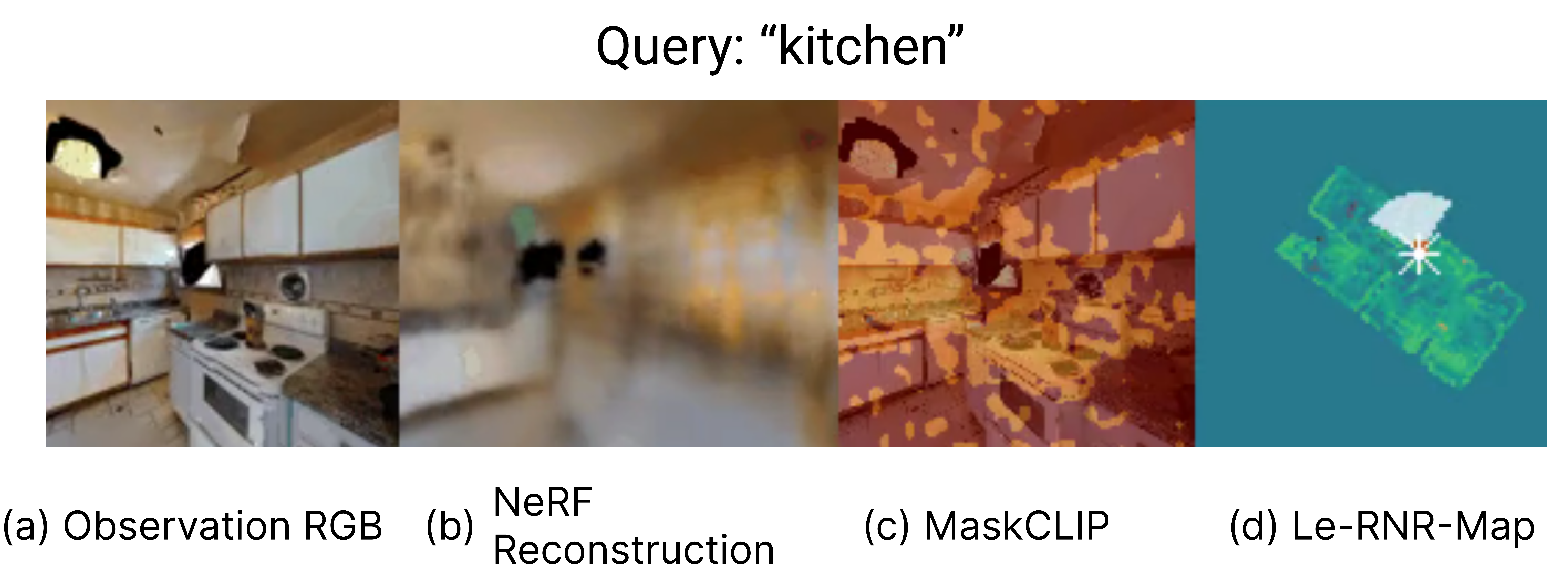}
    \caption{(a) Observation from Habitat-sim. (b) Reconstruction using latent code from \ours using Neural Radiance Field (c) Feature visualization of the text query using MaskCLIP. (d) Top-down view of \ours }
    \label{fig:similarity_maps_sup}
\end{figure}

In Fig. \ref{fig:similarity_maps_sup} we show an example of the final result of a search using \ours.
For video examples, we refer the reader to the video attached.

\section{Multi-Object Search}
\ours can also be used to look for multiple items. When given a set $Q$ of query prompts (e.g. \texttt{chair, couch, cabinet}), \ours can be explored to look for each desired item. Given a prompt in the form of ``$item_1, item_2, ..., item_n$'', we separate the single items by splitting the string on each comma. We then look for each item in sequence, following the standard procedure presented in the main paper.
We refer the reader to the video attached for some examples.
\section{Affordance Search}
In the following, we paste the code to compute the ``Affordance search'' using the available OpenAI chat-completion API, selecting the \texttt{gpt-3.5-turbo} model.
\begin{lstlisting}[language=Python]
import os
import openai
openai.api_key = "YOUR-API-KEY"

query = "Give me a location that can be relaxing after a long day at work"
completion = openai.ChatCompletion.create(
  model="gpt-3.5-turbo",
  messages=[
    {"role": "system", "content": f"You have the goal to act as an affordancy query mapper. I enter a query, and you have to give me A COMMA SEPARTED LIST of possible zone, inside the house, that satisfy my query. The ouput must be a sequence of location, without explantion. QUERY: {query}"}
  ]
)


possible_locations = completion.choices[0].message
multi_object_search(possible_locations)
\end{lstlisting}
Interesting future work will be devoted to different Large Language Models and prompts, and their impact on the final results. 

\subsection{Examples}
\textbf{query = ``Find me a drink to wake me up"}
\begin{lstlisting}[language=Python]
{
  "role": "assistant",
  "content": "kitchen, dining room, living room, office"
}
\end{lstlisting}

\textbf{query = ``Where can I wash my hands"}
\begin{lstlisting}[language=Python]
{
  "role": "assistant",
  "content": "bathroom, kitchen, utility room"
}
\end{lstlisting}

\textbf{query = ``Where can I watch the tv?"}
\begin{lstlisting}[language=Python]
{
  "role": "assistant",
  "content": "living room, bedroom, basement, media room"
}
\end{lstlisting}

\end{document}